\documentclass[10pt,twocolumn,a4paper]{esaAI}
\usepackage{subcaption}
\usepackage{placeins}
\usepackage{xcolor} 
\definecolor{customBlue}{RGB}{31, 48, 150}

\newcommand{\todo}[1]{\iffalse #1 \fi}

\title{XAMI - A Benchmark Dataset for Artefact Detection in XMM-Newton Optical Images}

\author[1]{Elisabeta-Iulia Dima\thanks{Corresponding author. Email: iuliaelisa15@yahoo.com}}
\author[2]{Pablo G\'{o}mez}
\author[2]{Sandor Kruk}
\author[2]{Peter Kretschmar}
\author[3]{\\Simon Rosen}
\author[1]{Călin-Adrian Popa}

\affil[1]{Department of Computers and Information Technology, Politehnica University of Timi\c{s}oara, Blvd. V. P\^arvan, No. 2, 300223 Timi\c{s}oara, Romania}

\affil[2]{European Space Agency (ESA), European Space Astronomy Centre (ESAC), Camino Bajo del Castillo s/n, 28692 Villanueva de la Cañada, Madrid, Spain}
\affil[3]{Serco Ltd., ESAC, Camino Bajo del Castillo s/n, 28692 Villanueva de la Cañada, Madrid, Spain}

\begin{document}

\makeCustomtitle

\begin{abstract}
Reflected or scattered light produce artefacts in astronomical observations that can negatively impact the scientific study. Hence, automated detection of these artefacts is highly beneficial, especially with the increasing amounts of data gathered. Machine learning methods are well-suited to this problem, but currently there is a lack of annotated data to train such approaches to detect artefacts in astronomical observations.
In this work, we present a dataset of images from the XMM-Newton space telescope Optical Monitoring camera showing different types of artefacts. We hand-annotated a sample of 1000 images with artefacts which we use to train automated ML methods. We further demonstrate techniques tailored for accurate detection and masking of artefacts using instance segmentation. We adopt a hybrid approach, combining knowledge from both convolutional neural networks (CNNs) and transformer-based models and use their advantages in segmentation. 

The presented method and dataset will advance artefact detection in astronomical observations by providing a reproducible baseline. All code and data are made available publicly\footnote{\url{https://github.com/ESA-Datalabs/XAMI-model}}$^,$\footnote{\url{https://github.com/ESA-Datalabs/XAMI-dataset}}. 


\end{abstract}



\section{Introduction}
\begin{figure}[!htb]
    \centering
    \includegraphics[width=0.9\linewidth]{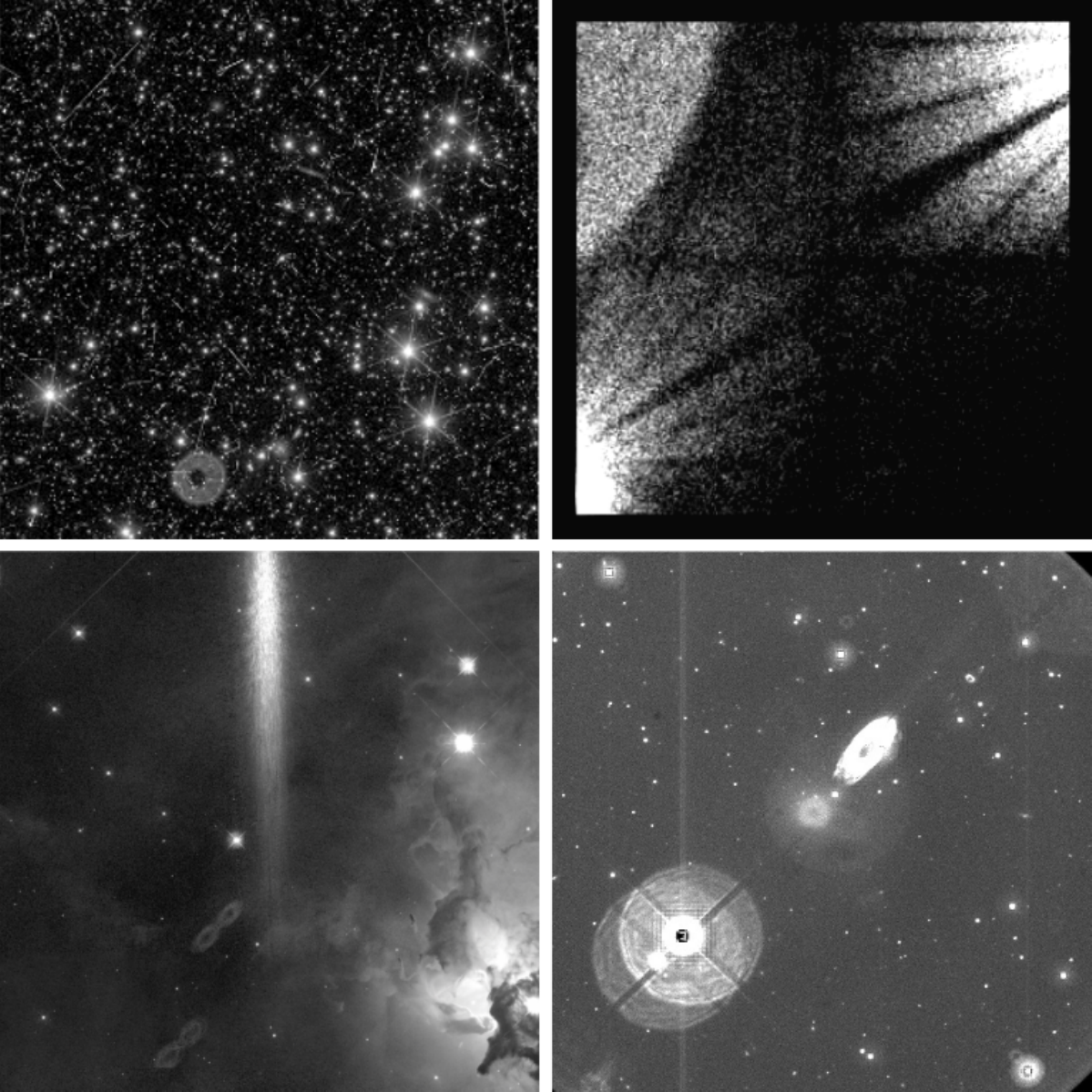}
	\caption{Examples of artefacts in various space missions. \textbf{(upper left)} An optical \textit{ghost} detected in Euclid’s First Light near-infrared images. \textbf{(upper right)} \textit{Ghost} rays and stray light patterns present in NuSTAR mission.  \textbf{(bottom left)} Star loops and \textit{dragon's breath} artefacts appearing in the Hubble Space Telescope images. \textbf{(bottom right)} Star loops and streaks present in the XMM-Newton Optical Monitor.}
\label{fig:artefacts_other_missions}
\end{figure}

 Astronomical surveys and space missions (e.g., LSST \cite{Ivezi__2019} and European Space Agency's Euclid mission \cite{laureijs2009euclid}) will enhance our understanding of the cosmos by delivering unprecedented images, measurements and insights into billions of stars and galaxies, the expansion of the Universe, dark energy and dark matter.  Such surveys will produce enormous amounts of data daily, thus the ongoing demand for the effective processing and analysis of large image data produced by space missions underscores the necessity for automated methodologies. The presence of artefacts (e.g. \textit{ghost} reflections, star loops, read-out-streaks) (e.g., \cref{fig:artefacts_other_missions}) poses challenges, potentially leading to false detections or affecting the photometric measurements of genuine sources.

 
\textbf{XMM-Newton Optical Monitor}. ESA's X-ray Multi-Mirror Mission (XMM-Newton) \cite{xmm_newton_2000, Schartel_2022} is an orbiting observatory with the principal goal to conduct detailed X-ray spectroscopy of various celestial objects. The XMM-Newton Optical Monitor (XMM-OM) \cite{Mason_2001, Cordova1989, Lumb1991} extends the simultaneous observational capability of the three main X-ray telescopes into the ultraviolet and optical bands. The XMM-OM source catalogue is a valuable resource containing approximately 9 million detections of around 6 million distinct sources. It plays a pivotal role in individual object analyses \cite{Soria_2001, refId0, 10.1111/j.1365-2966.2004.07660.x} and contributes significantly to survey science. However, the process of source detection within the XMM-OM data analysis process would benefit significantly from improved artefact recognition. 

Current non-AI approaches to detecting artefacts \cite{Mukhin_2023, article_nustar_straycats, DESAI201667} often struggle due to their reliance on generalised physical models. These models, while broadly applicable, fail to address specific scenarios effectively, leading to limitations in their practical utility. 

AI methods based on CNN and Vision Transformer (ViT) models have achieved notable success and have benefited real-world applications in tasks such as object detection \cite{wang2022yolov7, 10.1007/978-3-031-20053-3_27, maaz2022classagnostic, zong2023detrs} and segmentation \cite{srivastava2023omnivec, wang2022image, hümmer2023vltseg, https://doi.org/10.48550/arxiv.2401.15741, fang2022eva, wang2023internimage, liu2021swin, he2018mask, rs13234779}. Instance segmentation techniques for astronomical sources present significant progress \cite{10.1093/mnras/stad2785, Sortino_2023, hausen2022partialattribution}, yet there has been limited focus on artefacts detection \cite{tanoglidis2021deepghostbusters}. ViT models are increasingly preferred in computer vision due to their self-attention mechanisms. The Segment Anything Model (SAM) \cite{kirillov2023segment}, a ViT-based architecture, excels in class-agnostic instance segmentation and zero-shot learning, allowing it to identify objects not seen during training.

 We introduce \textbf{XAMI} (\textit{XMM-Newton optical Artefact Mapping for astronomical Instance segmentation}), a hybrid CNN and ViT-based model, and \textbf{XAMI-Dataset}, a high-precision instance segmentation dataset for astronomical images. Together, they provide a first baseline demonstrating ML-based artefact detection on astronomical images as well as benchmark and starting point for other researchers to build on.

\section{Methods}

\subsection{Dataset} 
We use 1000 single-channel images at various wavelengths (see \cref{table:dataset_info} and \cite{xmmom_filters_handbook}) from the XMM-OM as the baseline artefacts dataset. Each image comprises a stack of all available windows in a given filter of an observation that, together, cover the full $17'\times17'$ field of view. This corresponds to a full frame of $2048\times2048$ $\text{px}$ resolution, with an effective resolution of $0.477 '$/pixel. We rebinned the full-frame images to $512\times512$ $\text{px}$ for computational efficiency. We normalised images using \textit{ZScaleInterval} algorithm and enhanced them with \textit{Asinh} stretching to increase dynamic range without negatively affecting contrast. 

\begin{figure}[!htb]
    \centering
    \includegraphics[width=0.95\linewidth]{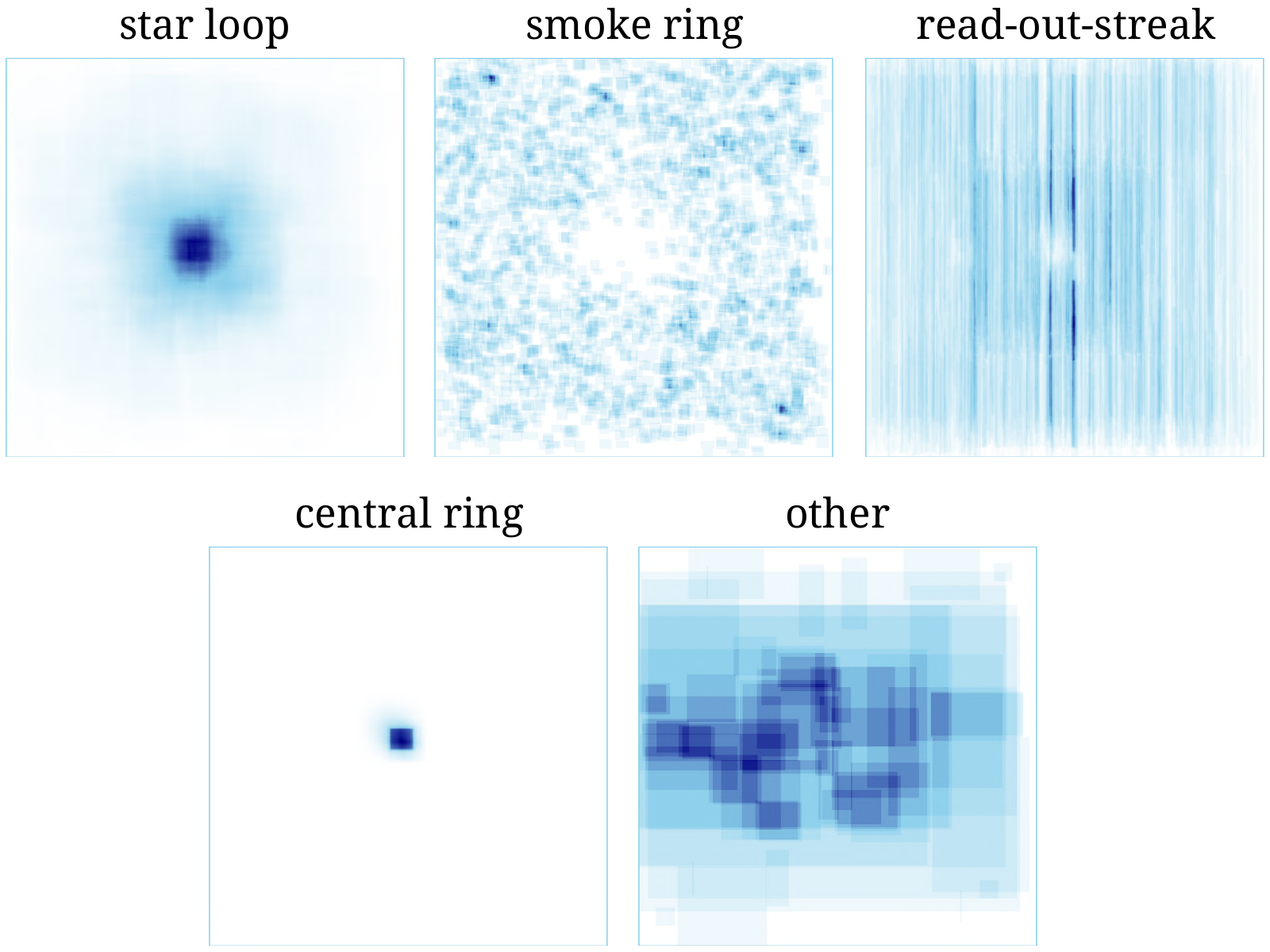}
	\caption{Distribution of annotation bounding boxes across different classes in the XAMI dataset.}
 
\label{fig:all_masks_distrib}
\end{figure}

The XAMI dataset consists of 7021 annotated artefacts which can be divided into the following categories (\cref{fig:all_masks_distrib}):

\begin{enumerate}
\item \textit{Read-Out-Streaks (ROS)} - arising from shutterless camera and continuous Charge-Coupled Device (CCD) photon recording during readout. 
\item \textit{Smoke rings (SR)} - resulting from internal reflections of starlight within the detector.
\item \textit{Central ring (CR)} - appearing in the centre of the detector, approximately $2'$ in diameter, resulting from background light scattering from a chamfer on the detector window mounting ring.
\item \textit{Star loops 
(SL)} - elongated scattered light features caused by light from bright stars within a $12'-15'$ off-axis range,  scattered from the chamfer.
\item \textit{Other} - other types of artefacts which usually represent scattered light spread over large areas. 

\end{enumerate}

\begin{table}[h]\renewcommand{\arraystretch}{1.0}
\begin{center}
\begin{tabular}{ccccc} 
\hline
\textbf{Filter} & $\lambda$(nm) & width & $\#$images & $\#$masks \\
\hline
V & 543 & 70 & 102 & 880\\
B & 450 & 105 & 116 & 1259 \\
U & 344 & 84  & 193 & 1837 \\
UVW1(L) & 291 & 83  & 403 & 2127\\
UVM2(M) & 231 & 48 & 175 & 681 \\
UVW2(S) & 212 & 50  & 63 & 226 \\
White(W) & 406 & 347 & 3 &	11 \\
\hline
\end{tabular}
\caption{Dataset information per observing filter, together with their central wavelength and width (nm).}
\label{table:dataset_info}
\end{center}
\end{table}

\begin{figure}[!htb]
    \centering
    \includegraphics[width=\linewidth]{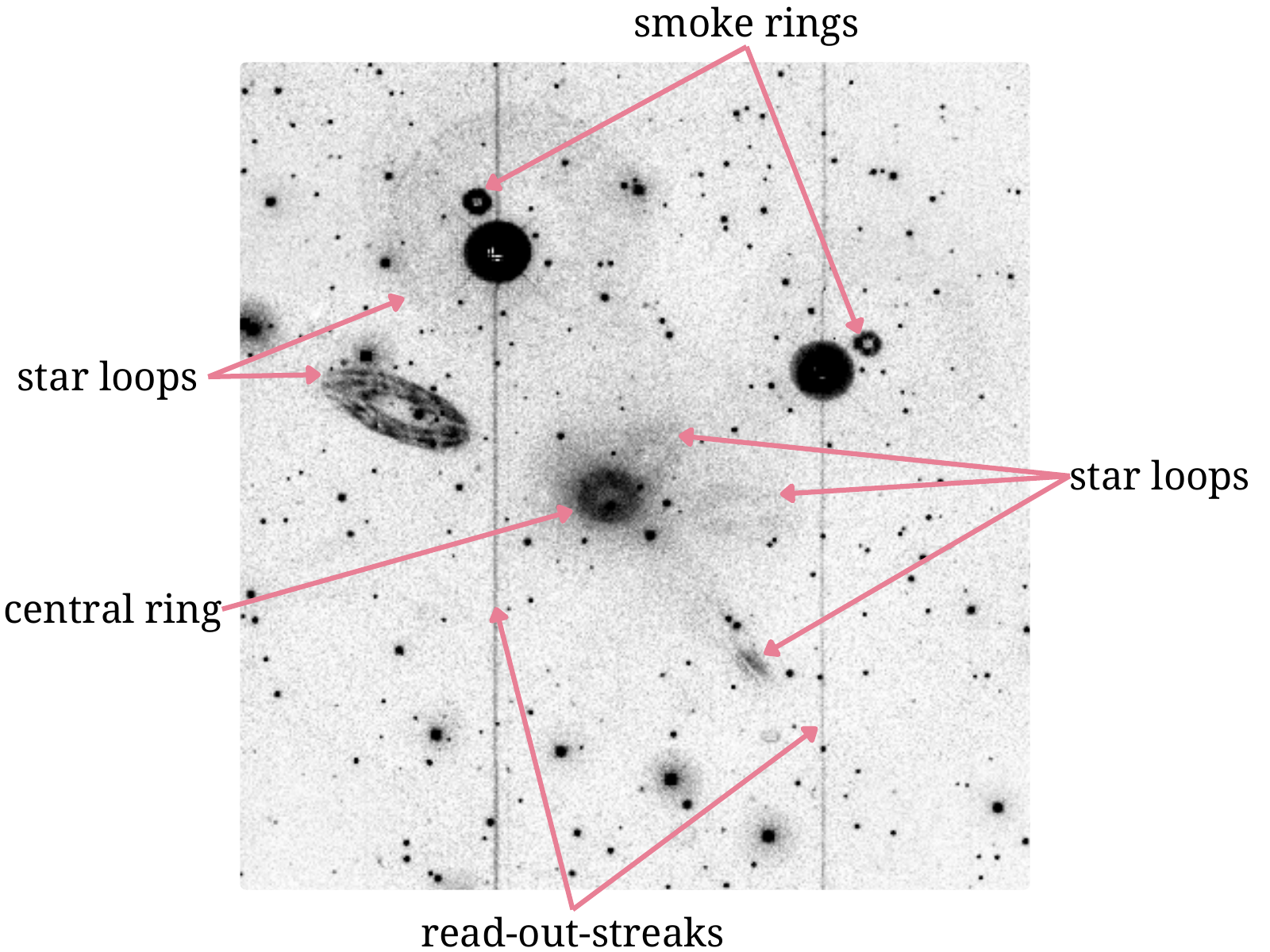}
	\caption{Artefacts appearing in the XMM-OM observation S0148740701 of the QSO 1939+7000 field (U filter).}
\label{fig:examples_artefacts}
\end{figure}

\begin{table}[h]\renewcommand{\arraystretch}{1.0}
\begin{center}
\begin{tabular}{lcc} 
\hline
\textbf{Class} & Train & Validation \\
\hline
CR   & 500 (9.43\%) & 168 (9.75\%)  \\
SR  & 1267 (23.91\%) & 402 (23.33\%)  \\
SL  & 1377 (25.99\%) & 467 (27.10\%)  \\
ROS  & 2122 (40.05\%) & 677 (39.29\%)  \\
Other  & 32 (0.60\%) & 9 (0.52\%)  \\
\hline
\end{tabular}
\caption{Dataset distribution across splits, given class labels. }
\label{table:dataset_distrib}
\end{center}
\end{table}

We use the stratified k-fold technique to maintain consistent class proportions across dataset splits, thus ensuring accurate performance estimation. Resulting class distributions can be seen in \cref{table:dataset_distrib}.

\subsection{Baseline Model}

We propose a class-aware approach for instance segmentation that integrates an object detector, specifically the YOLOv8 model \cite{reis2023realtime}, into our SAM prediction logic to facilitate auto-generated input prompts. 

Unlike CNNs, which strictly delineate object masks by bounding boxes, transformer-based models like SAM integrate self-attention to potentially extend beyond these initial margins. However, spatial invariance and accurate segmentation of faint objects remain a challenge for ViTs, in contrast with CNN approaches. By utilising SAM for smooth masks and YOLOv8 for faint objects with certain classes, we aim to overcome these limitations.


\section{Results}

\begin{figure}[!t]
   \centering
     \includegraphics[width=.99\linewidth]{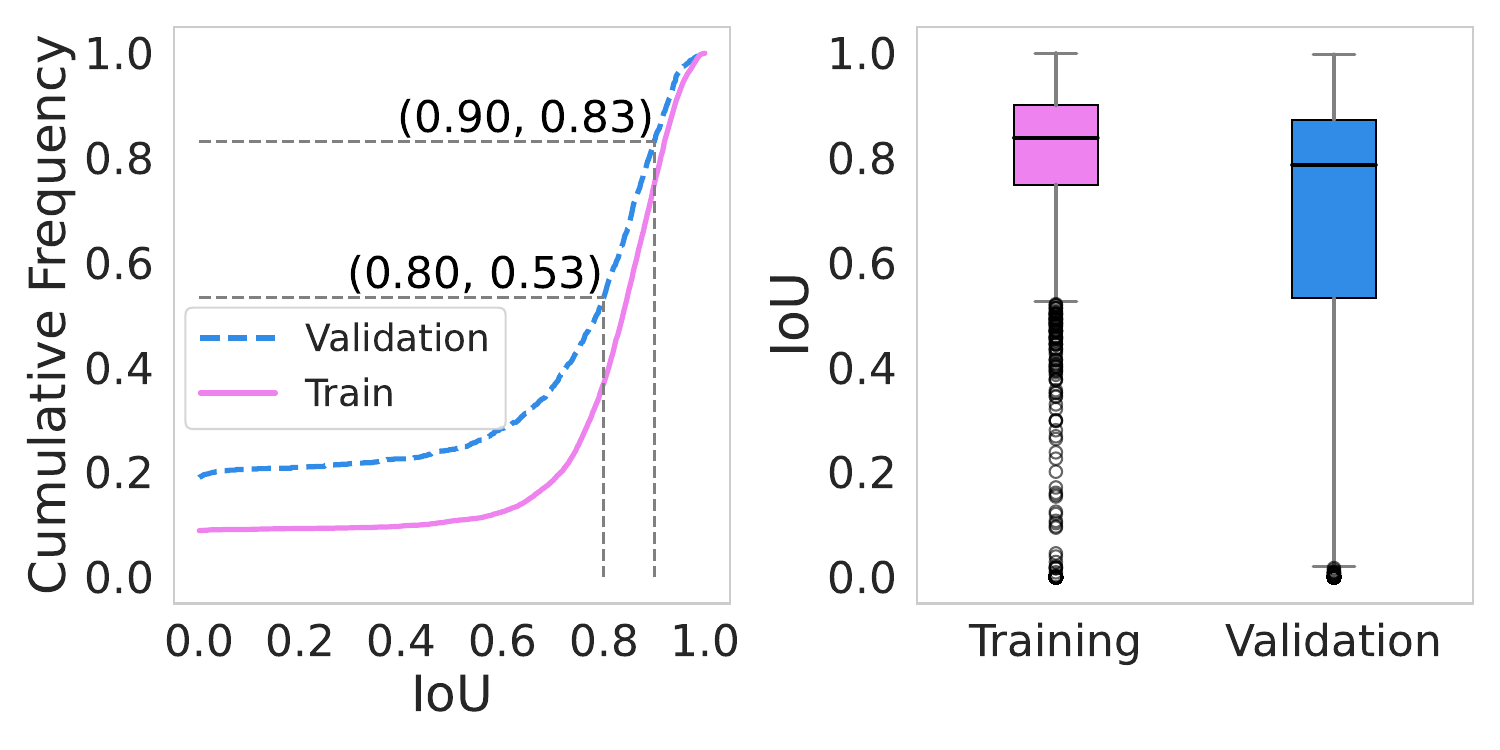}
   \caption{\textbf{(left)} Cumulative distribution of IoUs between predicted and true masks on training and validation sets. \textbf{(right)} Comparison of IoU distributions with higher median and consistency in training data and greater variability in validation data.}
   \label{fig:combined_ious_plot}
\end{figure}


\begin{figure*}[!t]
   \centering
     \includegraphics[width=.99\linewidth]{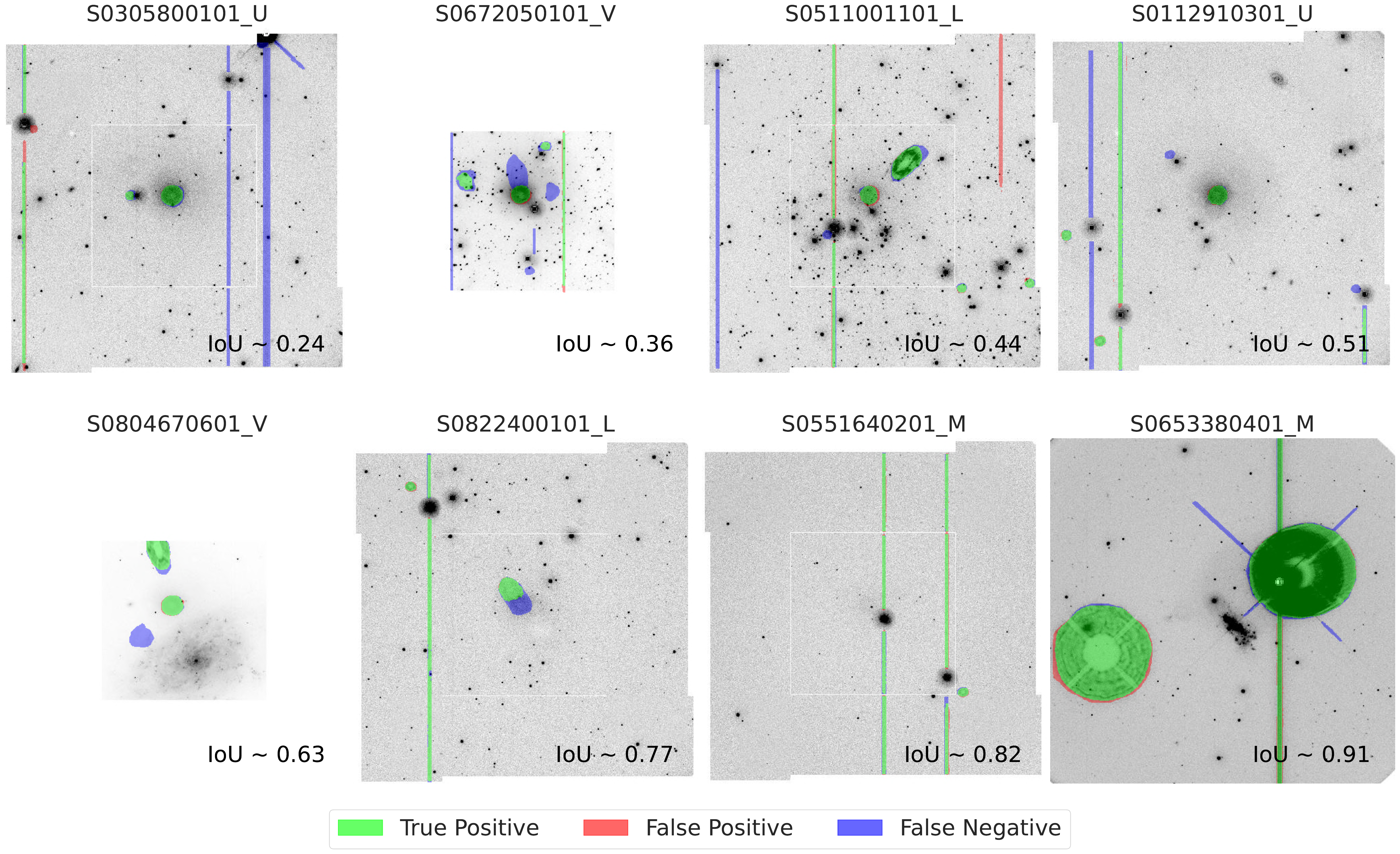}
   \caption{Detected masks across eight fields within the validation set, with increasing mean IoU between predicted and ground-truth masks. The mean IoU on the validation set images is $0.658\pm0.207$.
}
   \label{fig:ious_preds_tp_fp_tn}
\end{figure*}

Our methodology initially involves training SAM with ground-truth annotations using a distilled image encoder from MobileSAM \cite{zhang2023faster}. For SAM, images are resized to $1024 \times 1024$ $\text{px}$ and have their colours normalized. We use a batch size of 8, a warmup learning rate scheduler ($\mathrm{lr}_\mathrm{init}=3\times10^{-4}$, $\mathrm{lr}_\mathrm{final}=6\times10^{-5}$) for 16 steps, weight decay of $10^{-5}$ and AdamW optimizer. We train the Mask Decoder only, while freezing the Image Encoder and Prompt Embedding layers.

Following recommendations in \cite{kirillov2023segment}, we utilise the focal loss and dice loss in a 20:1 weighted scheme. At this stage, predicted and actual masks can be directly compared. Unlike usual SAM implementations, we choose to train the Intersection-over-Union (IoU) head to provide more representative segmentation metrics. Also, when generating masks, we configure the model to allow three predicted mask outputs and select the final mask based on the highest IoU score. The total loss integrates both segmentation and IoU loss.

After training the YOLO and SAM models separately to optimise their individual performances, we freeze the YOLO layers, couple its predicted bounding boxes to the SAM Prompt Encoder and continue training the SAM Mask Decoder to refine the segmentation process for 10 additional epochs. The alignment of predicted and ground truth masks is managed using the Kuhn-Munkres assignment algorithm \cite{https://doi.org/10.1002/nav.3800020109} by minimizing the IoU cost matrix. Due to higher spatial complexity of certain classes, particularly SL and \textit{Other}, we select YOLO masks for \textit{faint} objects of such classes at $1\sigma$ background level, as these predictions are more stable for low-intensity artefacts. We provide the precision and recall metrics (see \cref{table:prec-recall}) using a fixed seed for reproducibility. The formulas for precision and recall are defined as follows:

\begin{equation}
\text{P} = \frac{\text{TP}}{\text{TP} + \text{FP}} \quad \text{and} \quad \text{R} = \frac{\text{TP}}{\text{TP} + \text{FN}}
\label{eq:prec-recall}
\end{equation}

where TP - True Positive, FP - False Positive and FN - False Negative predictions.
\begin{table}[!h]\renewcommand{\arraystretch}{1.2}
\begin{center}
\begin{tabular}{lcc} 
\hline
 Category & Precision & Recall\\
\hline
Overall & 84.3 & 72.1 \\
CR & 89.3 & 94.0 \\
SR & 80.6 & 85.6 \\
SL & 80.5 & 74.1 \\
ROS & 71.1 & 73.3 \\
Other & 100.0 & 33.3 \\

\hline
\end{tabular}
\caption{Precision and recall resulted from model predictions on validation set. While smaller recall for \textit{Other} class may be caused by its under-representation, the CR class shows best overall performance, which may be attributed to its predictable location.}
\label{table:prec-recall}
\end{center}
\end{table}

\section{Discussion}
In our study, we enhanced artefact detection and segmentation in XMM-Newton images by integrating CNNs with ViT-based models, significantly boosting accuracy and reducing false positives in astronomical analysis. We combined traditional YOLO models for bounding box predictions with advanced SAM models for zero-shot segmentation, demonstrating the benefits of diverse neural network strategies in addressing complex image processing challenges. Despite improvements, high variation in exposure times and intensity levels in space imagery necessitate further model refinement tailored for astronomical missions. Additionally, these variations pose challenges in dataset annotation, eventually making it difficult to establish clear thresholds for distinguishing artefacts from the background. The XAMI average end-to-end inference time per image containing annotations is $100\text{ms}$, suitable for medium to large image data (up until hundreds of thousands of images similar to ours) and applications which do not particularly require instant real-time processing. The SAM heavy architecture still represents a bottleneck for prediction, with $70-80 \text{ms}/$image. 
We plan to expand our dataset with additional observations from various space missions to enhance model performance. Our goal is to benchmark our methods against other techniques in artefact detection and segmentation, while allowing the flexibility to choose or replace detectors and segmentors with novel state-of-the-art models. We will explore practical applications, such as faster data processing for space missions, and integrate our implementations into existing astronomical data processing systems. Feedback from astronomers will be crucial in further refining our approach.

\textbf{Acknowledgements}. The authors acknowledge the contribution of In\`{e}s Perez, Léa Zuili, Simon Astarita to dataset annotations. This publication uses data generated via the Roboflow.com\footnote{\url{https://app.roboflow.com/iuliaelisa/xmm_om_artefacts_512/}} and Zooniverse.org\footnote{\url{https://www.zooniverse.org/projects/ori-j/ai-for-artefacts-in-sky-images}} platforms.

\FloatBarrier

\printbibliography
\addcontentsline{toc}{section}{References}

\end{document}